\documentclass[letterpaper]{article} % DO NOT CHANGE THIS
\usepackage{aaai25}  % DO NOT CHANGE THIS
\usepackage{times}  % DO NOT CHANGE THIS
\usepackage{helvet}  % DO NOT CHANGE THIS
\usepackage{courier}  % DO NOT CHANGE THIS
\usepackage[hyphens]{url}  % DO NOT CHANGE THIS
\usepackage{graphicx} % DO NOT CHANGE THIS
\usepackage{natbib}  % DO NOT CHANGE THIS AND DO NOT ADD ANY OPTIONS TO IT
\usepackage{caption} % DO NOT CHANGE THIS AND DO NOT ADD ANY OPTIONS TO IT
\usepackage{algorithm}
\usepackage{algorithmic}

\usepackage{newfloat}
\usepackage{listings}

\usepackage{multirow}
\usepackage{subcaption}
\usepackage{microtype}
\usepackage{booktabs}
\usepackage[export]{adjustbox}
\usepackage{array}
\usepackage{bm}
\usepackage{todonotes}

\usepackage{amsmath}
\usepackage{tabularx}
\usepackage{enumitem}
\usepackage{marvosym}

\DeclareCaptionStyle{ruled}{labelfont=normalfont,labelsep=colon,strut=off} % DO NOT CHANGE THIS
\floatstyle{ruled}
\newfloat{listing}{tb}{lst}{}
\floatname{listing}{Listing}
\title{Enhancing Elusive Clues in Knowledge Learning by Contrasting Attention of Language Models}
\author{
    Jian Gao\textsuperscript{\rm 2}, 
    Xiao Zhang\textsuperscript{\rm 1}, 
    Miao Li\textsuperscript{\rm 1,\Letter}, 
    Ji Wu\textsuperscript{\rm 1,3,4}
    \\
}
\affiliations{
     \textsuperscript{\rm 1}Department of Electronic Engineering, Tsinghua University \\
     \textsuperscript{\rm 2}Department of Energy and Power Engineering, Tsinghua University \\
     \textsuperscript{\rm 3}College of AI, Tsinghua University \\
     \textsuperscript{\rm 4}Beijing National Research Center for Information Science and Technology\\
     \{gaojian21, xzhang19\}@mails.tsinghua.edu.cn\\
     \{miao-li, wuji\_ee\}@tsinghua.edu.cn   \ \ \
}

\begin{document}

\maketitle

\begin{abstract}
Causal language models acquire vast amount of knowledge from general text corpus during pretraining, but the efficiency of knowledge learning is known to be unsatisfactory, especially when learning from knowledge-dense and small-sized corpora.
The deficiency can come from long-distance dependencies which are hard to capture by language models, and overfitting to co-occurrence patterns and distracting clues in the training text.
To address these issues, the paper proposes a method to enhance knowledge learning during language model pretraining, by enhancing elusive but important clues in text discovered by the language model themselves.
We found that larger language models pay more attention to non-obvious but important clues, which are often overlooked by smaller language models. Therefore, we can identify these clues by contrasting the attention weights of large and small language models. We use the identified clues as a guide to perform token-dropout data augmentation on the training text, and observed a significant boost in both small and large models' performance in fact memorization. This shows that the behavior contrast between more and less-performant language models contains important clues for knowledge learning, and it can be ``amplified" for a straight-forward improvement in knowledge learning efficiency.

\end{abstract}

\begin{links}
\link{Code}{https://github.com/tsinghua-msiip/contrasting_attention}
\end{links}

\section{Introduction}

Pretrained large language models have shown impressive performance on a wide variety of downstream tasks \citep{instructgpt, flan, llama}. To achieve good generalization, these models need to be trained on web-scale corpora that are diverse and large enough to capture the complexity of natural language. Unfortunately, it is observed that when training corpora is limited in size or style variation, language models can struggle to generalize the information learned from the corpora \citep{allenzhu3.1}. This deficiency poses a challenge for injecting knowledge into pretrained language models via continual pretraining (finetuning). In many domains, the available corpora is often limited and knowledge-dense (e.g., in forms of textbooks, manuals, documentations). Such domain text may be difficult to be utilized effectively in finetuning, and the language models may not be able to effectively generalize the domain knowledge to downstream domain tasks.

% Attention mechanism and its role in model performance
Not very much is known about the causes of such deficiency in knowledge learning. One likely cause is overfitting to co-occurrence patterns in the limited training text, causing learning of spurious correlations instead of correct factual associations. Another possible reason is the difficulty of capturing long-range dependencies in text, which are crucial for understanding complex relationships. Such deficiency is sometimes a result of intentional design choice in the model architecture, such as the decay of attention weights in the RoPE \citep{rope} positional encodings.

One possible route to understanding this phenomenon is via the attention module in language models. The attention mechanism is a key component that allows the model to focus on different parts of the input when making predictions. The attention weights are shown to be interpretable and explaining the model's behaviors \citep{DBLP:journals/corr/abs-1906-04341}.

Recently, \citet{DBLP:journals/corr/abs-2309-15098} show that when predicting factual information, models are less likely to attend to the correct clue  if the model does not know about the fact. This implies that for new knowledge unknown to the model, the model may not be able to attend to the correct clue at first, leading to difficulty in associating the correct clue (e.g., the head entity) with the prediction target (the tail entity).

To help language models learn, especially smaller models, a common approach is to use knowledge distillation \citep{DBLP:journals/corr/HintonVD15} (or teacher-student method) to transfer knowledge from a larger model. Given a learning goal, a more performant language model such as GPT-4 \citep{gpt4} is often used to generate training data for the smaller model \citep{kdsurvey}. A main drawback of this approach is that it requires the larger model to be already capable of the task or already have the knowledge. This make it not suitable for learning novel knowledge, such as new facts from an evolving domain. Also, it can only help the smaller model to learn but cannot help the larger model.

In this paper, we propose a simple method to enhance factual knowledge learning in continual pretraining, with the help of a pair of larger and smaller models. Our method is effective in learning novel facts and can boost the performance of both the larger and smaller models. The main contributions of the paper are as follows:

\textbf{Attention difference between large and small language models reveals elusive but important clues in text.} We show that while large and small language models both show high attention to important and obvious clues in text, large models pay significantly more attention than smaller models to important clues that are less obvious or elusive. Therefore, by contrasting the attention weights of large and small models, we can identify these elusive clues in text that are important for knowledge learning but are often easily overlooked.

\textbf{Augmenting elusive clues in text boosts knowledge learning in continual pretraining.} We show that by using the identified elusive clues as a guide, a token-dropout data augmentation that highlights the elusive clues can significantly boost the model's performance in knowledge learning. We experimented on both synthetic and real-world corpus and show that the proposed method outperforms other forms of data augmentation, and boosting elusive clues universally helps both the large and the small models.

To the best of our knowledge, we are the first to analyze the the attention discrepancies between large and small models and use it for data augmentation. Prior work have distilled attention pattern from large models to small models, but without analyzing what is being distilled. Unlike distillation, our approach also enhances the performance of large models, which is a novel contribution on our part.

We release the code and data used in this paper for reproducibility and further research.

\section{Related Work}

\label{gen_inst}

\subsection{Attention as Behavior Explanation}

It is observed that attention weights in transformer models provide interpretable clues about the model's behavior. For example, attention heads within multi-head attention can spontaneously differentiate into distinct roles \citep{DBLP:journals/corr/abs-1906-04341}. Certain heads play a more significant role and affect performance significantly \citep{DBLP:journals/corr/abs-1905-09418}. More performant models tend to have attention weights that focus more on key information and features, a possible explanation of their superior performance \citep{DBLP:journals/corr/abs-2309-15098}.

Some argue that while attention is somewhat interpretable, its interpretability is not an indicator of model performance \citep{DBLP:conf/acl/SerranoS19}. There is divided opinion on the extent to which attention weights reflects true model behavior \citep{DBLP:conf/naacl/JainW19,DBLP:conf/emnlp/WiegreffeP19}. Our study extends these findings by comparing and contrasting attention weights of different models, and show that the difference between attention weights of large and small models can provide important behavioral clues.

\subsection{Data Augmentation on Text}

%%% ver 2
Data augmentation is a critical technique for enhancing robustness and generalization, especially for limited-size datasets. Various data augmentation methods have been proposed, including random editing of sentences \citep{DBLP:conf/emnlp/WeiZ19} such as insertion, swapping, and deletion. Synonym replacement methods \citep{DBLP:conf/aist/MosolovaFB18, DBLP:conf/cikm/RizosHS19} replace words with their synonyms. Contextual augmentation methods \citep{DBLP:conf/naacl/Kobayashi18} replace words with other words predicted by a language model for semantic variations. Back-translation \citep{backtranslation, backtranslation2} is another commonly used method that generates augmented data by translating to and then back from another language. More sophisticated methods combine multiple augmentations \citep{DBLP:conf/nips/XieDHL020, DBLP:conf/emnlp/KarimiR021}.

% att
Given that attention provides interpretable clues about the model's behavior, \citet{DBLP:journals/ict-express/YuYJK22, DBLP:conf/icmlc2/HailemariamLMA23} uses attention weights to find semantically significant words for replacement augmentation. \citet{DBLP:journals/corr/abs-2309-11104} uses attention weights to find significant input parts for mixup augmentation \citep{mixup}. We go a step further and show that only augmenting the most significant words is insufficient for challenging knowledge learning scenarios, and augmenting hard-to-notice but important parts of the input boosts the model's performance even better than augmenting the significant parts.

\subsection{Teacher-Student Methods for Language Models}

%%% ver 2
To enhance the performance of smaller models, knowledge distillation methods have been extensively developed to transfer knowledge from larger models to smaller models \citep{DBLP:journals/corr/HintonVD15, kdsurvey}. Large pretrained language models can be used to generate data for finetuning smaller models to transfer its knowledge and skills, for example, instruction following \citep{selfinstruct, vicuna} and reasoning ability \citep{reasondistill, distillreasoning2}. Distillation from large model is also frequently used to build strong domain or task-specific models with a compact size, like for coding \citep{phi1, codellama} and math \citep{wizardmath, mammoth}. Our work explores a different way to utilize large models: we find the behavior difference between large and small models and use it to guide the models towards more difficult part of the text.

\subsection{Continual Pretraining of Language Models}
Continual pretraining takes a language model pretrained on a general corpus and continual the pretraining process with a new corpus, typically domain-specific text, to enhance the model's performance on domain tasks. Model acquires new knowledge and ability via continual pretraining, for example, in coding \citep{codex}, math \citep{minerva}, and medicine \citep{medpalm2}. We aim at learning new factual knowledge from text via continual pretraining, similar to those in \citep{continuallmknowledge, allenzhu3.1}.

\section{Problem Setup: Knowledge Learning Deficiency}
\label{problem}

\subsection{Task: Fact Learning in (Continual) Pretraining}

% Introduce the synthetic biography dataset \citep{allenzhu3.1} with an example. 

Language models can learn factual knowledge from pretraining (or continual pretraining) on text corpora. \citet{allenzhu3.1} introduced a synthetic biography dataset for evaluating the efficiency of knowledge learning in language models. The dataset has been utilized by \citep{Khalifa2024SourceAwareTE}, \citep{Golovneva2024ReverseTT}, and \citep{Saito2024WhereIT}. It consists of short synthetic biographies of individuals, with a fixed format shown in the following example:

% {\leftskip=1cm\relax
%  \rightskip=1cm\relax
\begin{quote}
 \textit{\textbf{Liam Thompson} was born on \textbf{January 5, 1990}. He spent his early years in \textbf{Melbourne, Australia}. He received mentorship and guidance from faculty members at \textbf{Sorbonne University}. He completed his education with a focus on \textbf{Biomedical Engineering}. He had a professional role at \textbf{the British Museum}.}
\end{quote}
 % \par}

Each biography contains information about an individual's name, birth date, birth city, education, and job status. The task is to finetune (continual pretraining) a language model on the biographies to let it memorize the factual information about the individuals. After training, the model is evaluated on a question-answering task, where we evaluate the model's accuracy in memorizing the underlined part of the biographies.

The questions are formatted like ``\textit{When was Liam Thompson born?}". 
% Details on the training corpus and evaluation data are provided in the appendix.
When questions were rephrased using GPT-4, performance generally declined, indicating that the original questioning format yielded the best performance, so that question style
has minimal impact on our conclusion.

\subsection{Deficiency in Knowledge Learning Over Long-Range Dependency}

\citet{allenzhu3.1} have shown that training language models from scratch on the biographies yield poor performance in question answering. We instead perform continual pretraining on pretrained language models up to 70 billion parameters. The language models have undergone extensive pretraining on massive corpora and show strong language capabilities.

We show that even pretrained models with billions of parameters struggle to memorize facts perfectly in continual pretraining. Table \ref{tab:plain-finetuning} 
show that while Gemma 2 \citep{gemma2} and LLaMA 3 \citep{llama3} memorize the first two pieces of information (birth date and birth city) with high accuracy, they struggle to memorize the following three pieces of information (university, major, and company). This rules out the possibility that the performance deficiency is due to limited model size or insufficient pretraining.
We also tried swapping the positions of five kinds of information resulted in the same trend: accuracy decreases as distance increases, demonstrating that long-range dependencies, rather than en-tity types, are the primary cause of poor performance.

\begin{figure}[h]
\includegraphics[scale=0.37, center]{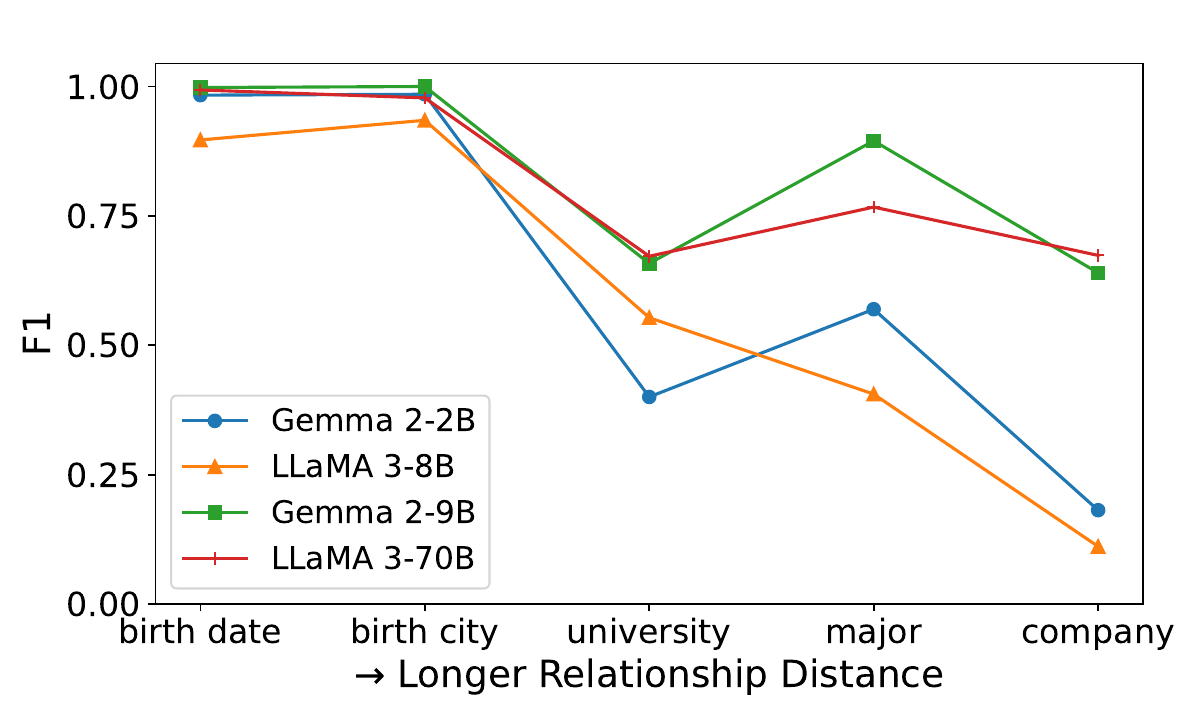}
\caption{Performance on the QA task show a decreasing trend as the distance between the head and tail entities in the relationship increases in the training text.}
\label{fig:perf_with_distance}
\end{figure}

The performance trend on QA tasks is also plotted in Figure \ref{fig:perf_with_distance}. It is clear that as the relationship spans longer distances (i.e., the distance between the tail entity, such as ``Company", to the head entity name, the person's name), the model's performance show a decreasing trend. This indicates that the model struggles to capture long-range dependencies in text, which is crucial for learning complex relationships.

\setlength{\tabcolsep}{0.3mm}

\begin{table}[h!]
\centering

\fontsize{7}{10}\selectfont
\begin{tabular}{lcccccc}
\toprule
& & \makebox[0.07\textwidth][c]{\textbf{\small Date }} & \makebox[0.04\textwidth][c]{\textbf{\small City }} & \textbf{\small University} & \textbf{\small Major} & \textbf{\small Company} \\
\midrule
\multirow{2}{*}{\textbf{\small LLaMA 3 8B}} & \small EM & \small 0.82 & \small 0.91 & \small 0.20 & \small 0.34 & \small 0.09 \\
& \small F1 & \small 0.90 & \small 0.93 & \small 0.55 & \small 0.41 & \small 0.11 \\
\midrule
\multirow{2}{*}{\textbf{\small LLaMA 3 70B}} & \small EM & \small 0.98 & \small 0.95 & \small 0.36 & \small 0.73 & \small 0.66 \\
 & \small F1 & \small 1.00 & \small 0.98 & \small 0.67 & \small 0.77 & \small 0.67 \\
\midrule
\multirow{2}{*}{\textbf{\small Gemma 2 2B}} & \small EM & \small 0.98 & \small 0.99 & \small 0.12 & \small 0.54 & \small 0.15 \\
& \small F1 & \small 0.98 & \small 0.99 & \small 0.40 & \small 0.57 & \small 0.18 \\
\midrule
\multirow{2}{*}{\textbf{\small Gemma 2 9B}} & \small EM & \small 0.99 & \small 1.00 & \small 0.51 & \small 0.89 & \small 0.63 \\
 & \small F1 & \small 1.00 & \small 1.00 & \small 0.66 & \small 0.90 & \small 0.64 \\
\bottomrule
\end{tabular}
\caption{Performance on the QA task after continual pretraining on the biography corpus.}
\label{tab:plain-finetuning}
\end{table}

One possible reason for the deficiency in learning long-range dependencies is overfitting to a large amount of distracting information between the head and tail entities in a relationship. Overfitting is more likely when relationship only occur in few examples like in the biography dataset. 
Another possible reason comes from the bias in the model architecture that biases the model's attention towards nearby information. Many popular models, such as LLaMA and Gemma, use the Rotary Position Embedding (RoPE) \citep{rope} as positional encoding in their attention module. RoPE has a long-term decay property, which means that attention weights decay as the relative distance between the key and value token increases. This makes the model focus more on adjacent information but at a cost of important information that are occasionally far-away, hurting the model's performance in learning long-range dependencies.

% A plot showing that accuracy decreases with distance.

\section{Analysis: Contrasting Attention of Language Models}
\label{analysis}

We have shown that language models could achieve near-perfect accuracy in memorizing relationships that span a short distance in text, but struggle when they span a longer distance. In this section, we use attention weights as an interpretability tool to analyze the model's behavior while learning long-range dependencies. We show that LLMs can pay inadequately little attention to key information that is located further away, and more performant larger models can pay more attention to these information than smaller models.

\subsection{Attention Weight Visualization}

We look at model's attention weights to try answering the following question: what information does the model pay attention to when predicting the tail entities in a relationship? The model uses attention weights to retrieve hidden states of context tokens, therefore the weights determines the information flow from the context to the current token in text. Furthermore, if an incorrect head entity is attended to when predicting the tail entity during the forward pass, in backpropagation the model will likely reinforce this incorrect association and cause the model to learn the wrong relationship.

\begin{figure*}[ht!]

\begin{subfigure}{0.49\textwidth}
\includegraphics[width=1\linewidth]{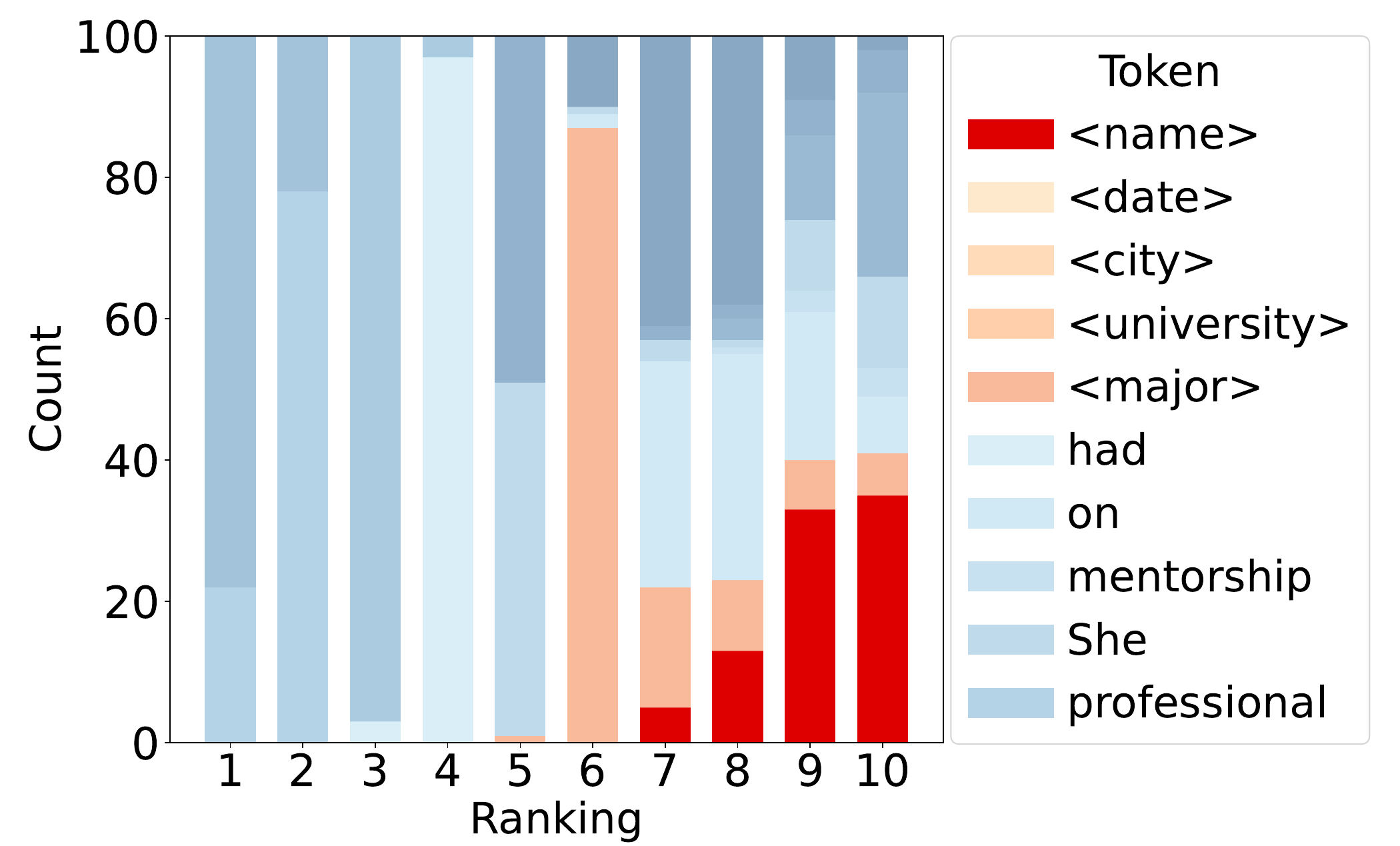} 
\captionsetup{skip=-0.3pt, singlelinecheck=false, margin={70pt,0pt}}
\caption{Gemma 2 2B}
\label{Gemma-2-2B}
\end{subfigure}
\begin{subfigure}{0.49\textwidth}
\includegraphics[width=1\linewidth]{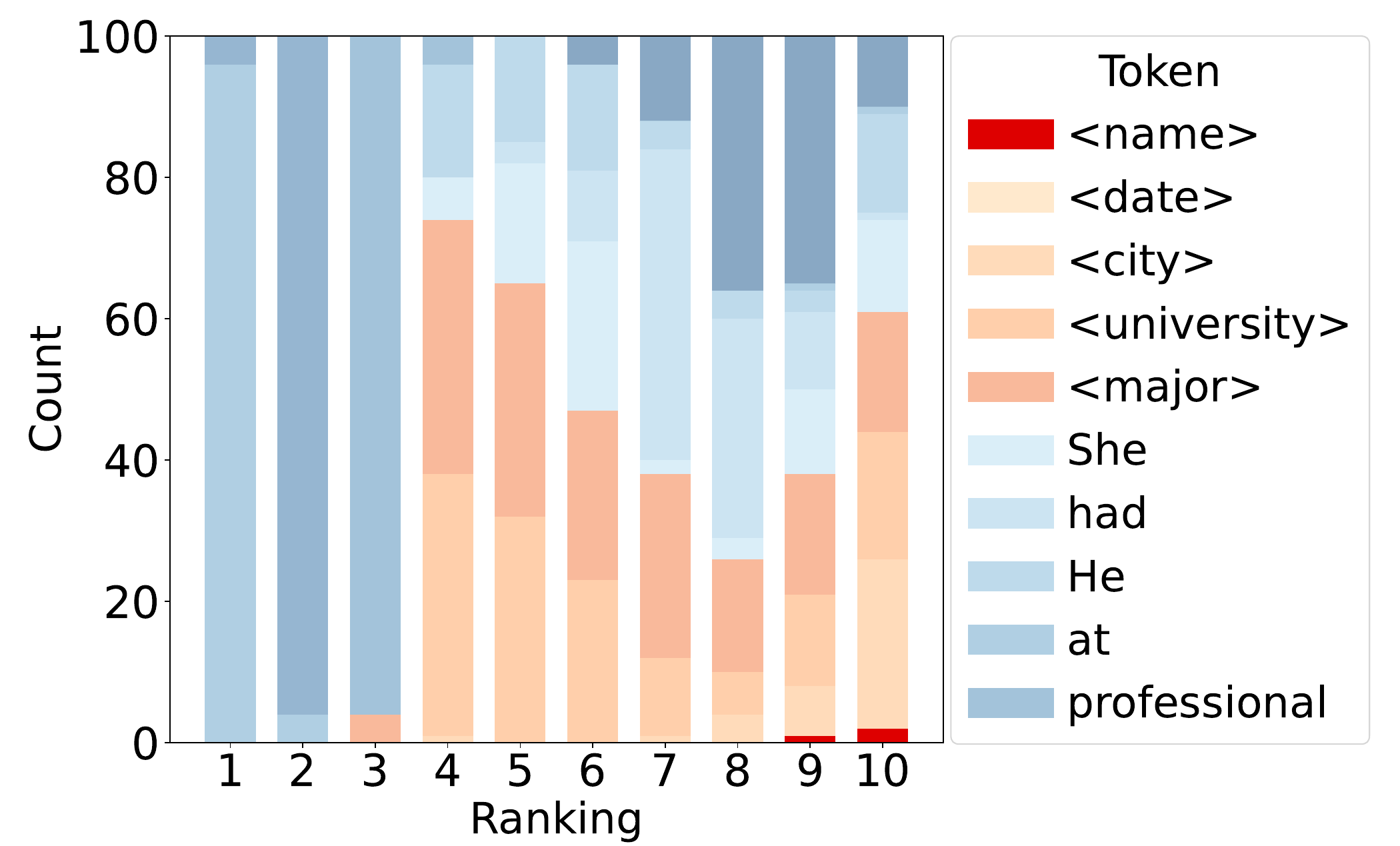} 
\captionsetup{skip=-0.3pt, singlelinecheck=false, margin={70pt,0pt}}
\caption{LLaMA 3 8B}
\label{LLaMA-3-8B}
\end{subfigure}

\begin{subfigure}{0.49\textwidth}
\includegraphics[width=1\linewidth]{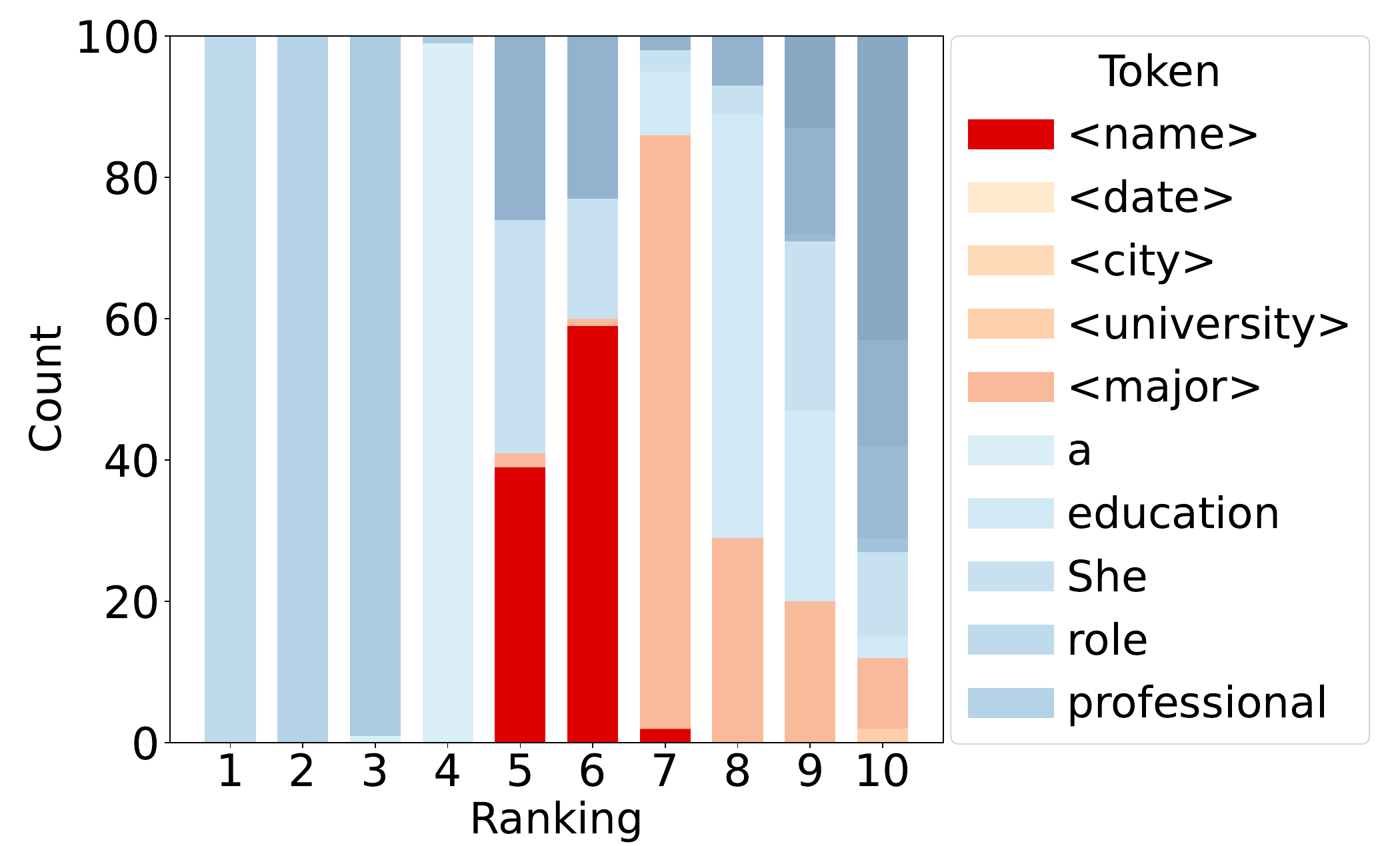}
\captionsetup{skip=-0.3pt, singlelinecheck=false, margin={70pt,0pt}}
\caption{Gemma 2 9B}
\label{Gemma-2-9B}
\end{subfigure}
\begin{subfigure}{0.49\textwidth}
\includegraphics[width=1\linewidth]{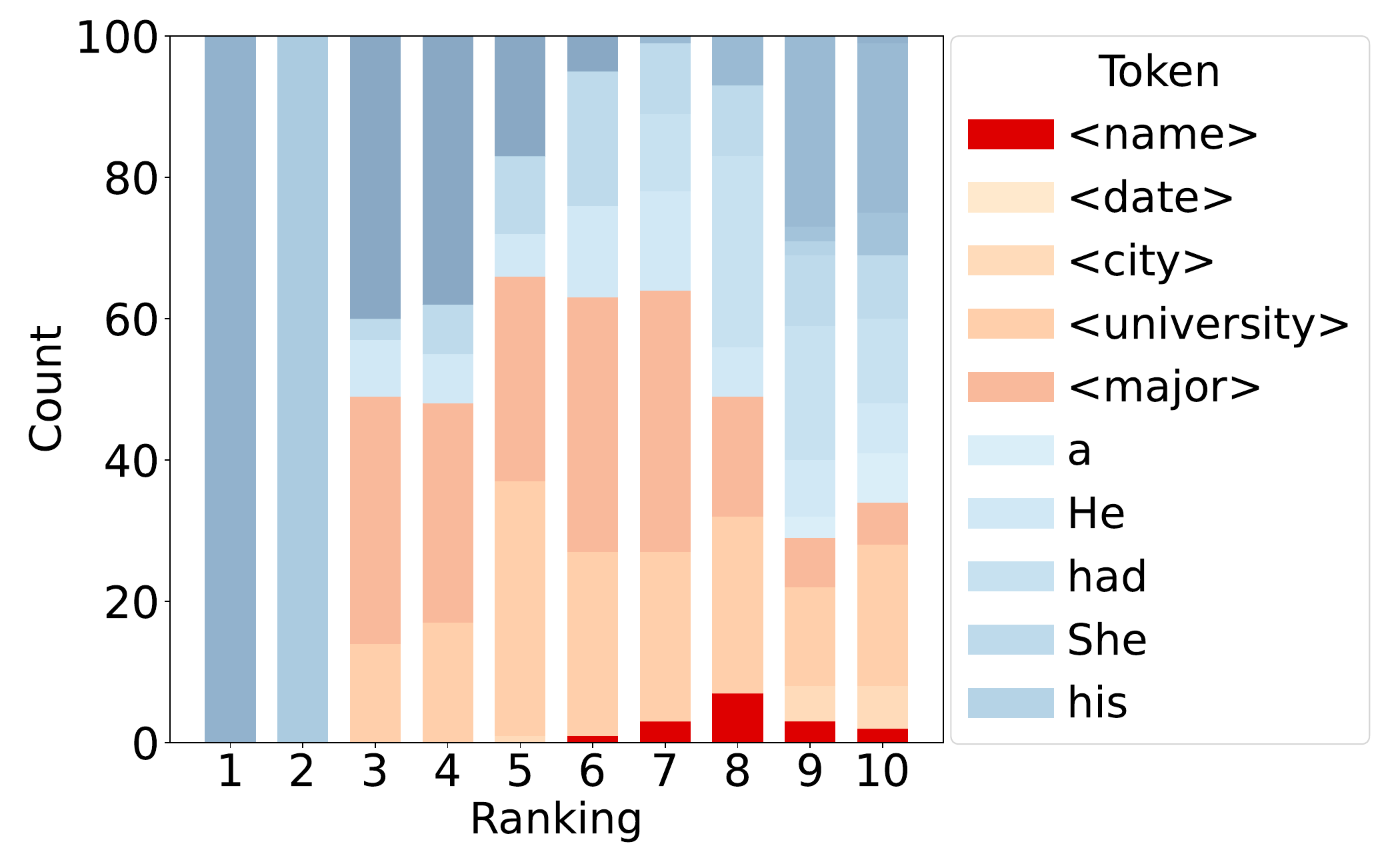}
\captionsetup{skip=-0.3pt, singlelinecheck=false, margin={70pt,0pt}}
\caption{LLaMA 3 70B}
\label{LLaMA-3-70B}
\end{subfigure}

\caption{Visualization of tokens receiving the highest attention weights, at the preposition just before the ``company" field. Tokens in a sentence are ranked by attention weight, from large to small. Each bar in the graph show the constitution of the i-th ranked token from 100 biographies. ``$\langle...\rangle$" denotes tokens belonging to the information fields, and all else are individual tokens.
Models generally pay most attention to the relationship words (e.g., ``professional", ``role", ``at"), then to distrating entities in between (e.g., birth date, city, etc.). 
Because LLaMA 3 models have no special start token at the front of sentences, we add "Text: " at the beginning of sentences to avoid impact of the special position of tokens. All visualization results of LLaMA 3 are done in this way.}
\label{fig:attention}
\end{figure*}

To visualize model's attention weights when predicting the tail entities in a relationship, we extract the attention weights at the \textit{preposition tokens}, i.e., the word immediately preceding the tail entity. For example, in the sentence ``He received mentorship and guidance from faculty members \textit{at} Sorbonne University", the attention from the token ``\textit{at}" is extracted. Because the model is predicting the tail entity ``Sorbonne University" at this position, the attention weights\footnote{To simplify analysis, we took the approach of averaging the attention weights across all layers and attention heads.} here likely corresponds to the  information necessary for predicting it. To ease visualization and for better comparison, instead of directly showing the attention weights, we rank the tokens and visualize the top 10 tokens with the highest attention weights. For each model, we calculate the token attention ranking for 100 biographies\footnote{Because attention paid on meaningless tokens provides little information, we removed periods, commas, spaces, and placeholders at the beginning of a sentence(for example, \textless bos\textgreater).}, and summarize the ranking using a bar plot in Figure \ref{fig:attention}.

% \begin{figure}[h]

Results show that models assign the most attention to the most important information for predicting the tail entity: the relationship words. The model also pays much attention to the distracting entities in the preceding text. The correct head entity, which is the key information for predicting the tail entity, receives hardly any  attention from smaller models and only a small amount of attention from larger models such as Gemma 2 9B, and is almost never ranked in top tokens.
This indicates that the model's attention is biased towards short-distance information, which may lead to the model learning the incorrect association and overfitting to such spurious co-occurrences.

% Visualize tokens by attention ranking.

\subsection{Contrasting Attention of Large and Small Language Models}

Comparing to smaller models, larger language models tend to have overall better language understanding capabilities, therefore could be more likely to pay attention to the correct clue in the text. For a same family of models, for example, the LLaMA 3 8B and 70B models, the training corpus, model architecture, and training procedure are mostly similar, and they should have relatively similar general behavior pattern besides their capability differences. 

\begin{figure*}[ht!]
\centering
\begin{subfigure}{0.49\textwidth}
\includegraphics[width=1\linewidth]{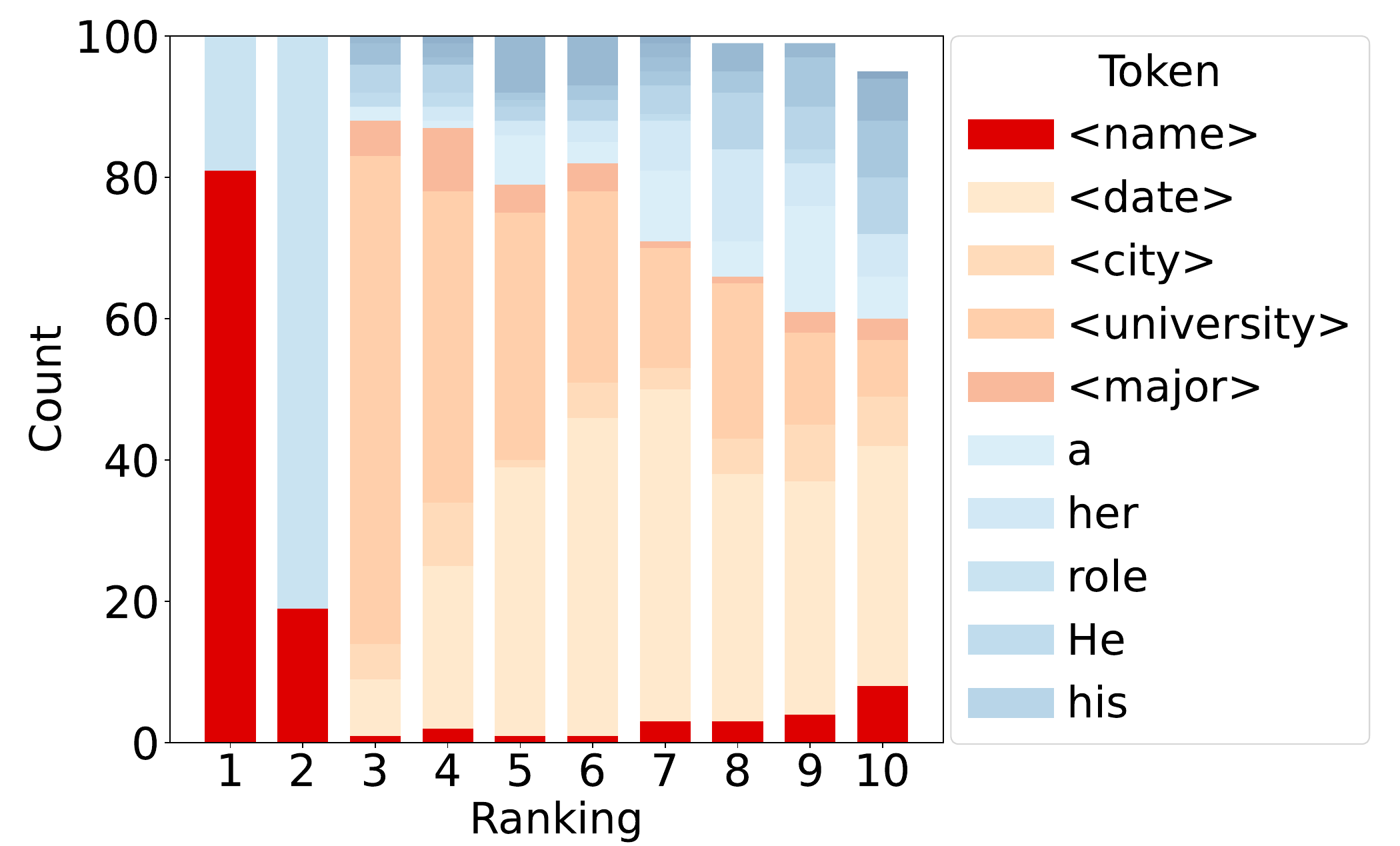} 
\captionsetup{singlelinecheck=false, margin={70pt,0pt}}
\caption{Gemma 2 9B$/$2B}
\label{gemma-7b-2b}
\end{subfigure}%
\begin{subfigure}{0.49\textwidth}
\includegraphics[width=1\linewidth]{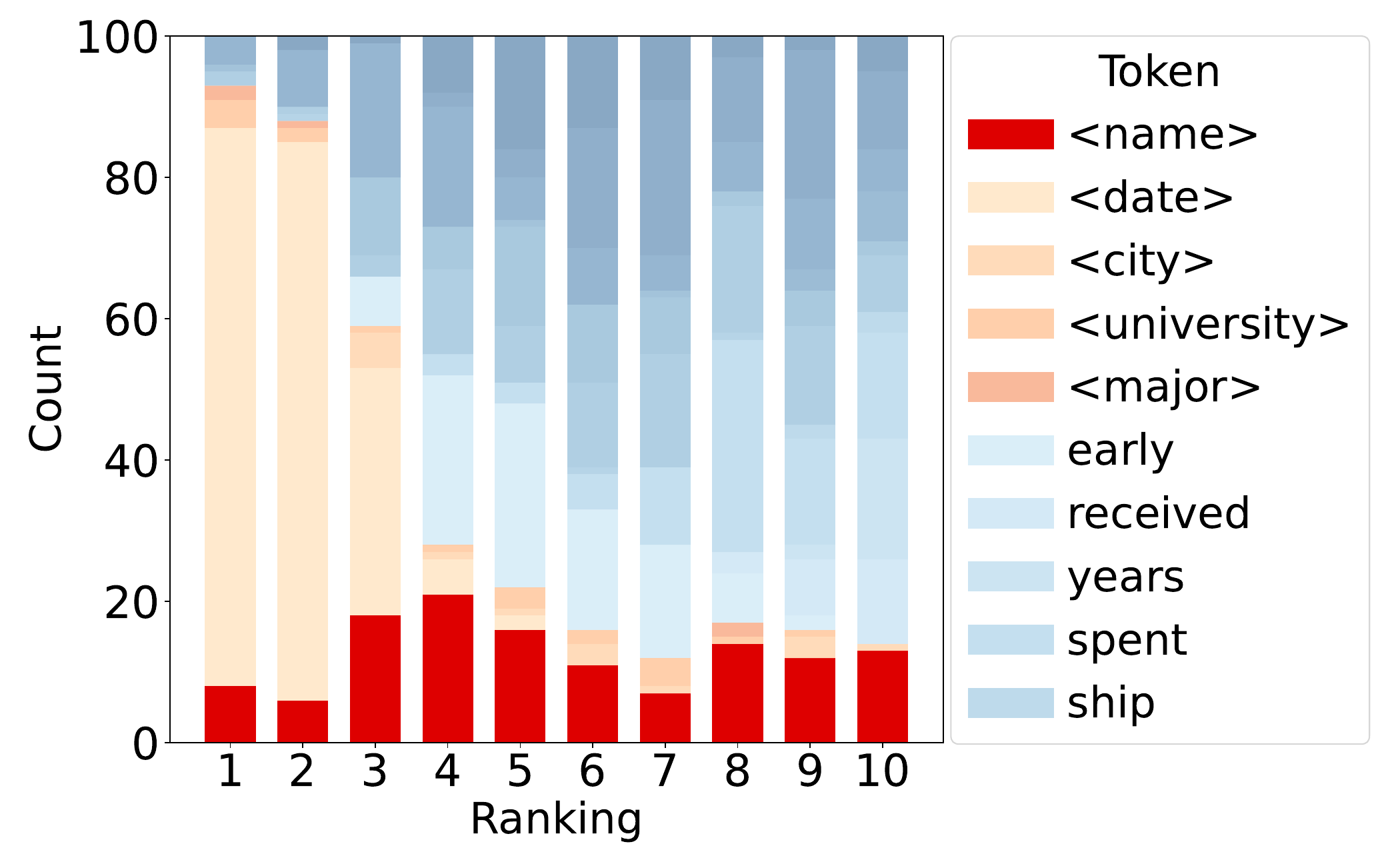}
\captionsetup{singlelinecheck=false, margin={70pt,0pt}}
\caption{LLaMA 3 70B$/$8B}
\label{llama2-70B-7B}
\end{subfigure}

\caption{Visualization of tokens receiving the highest additional attention weights from the large model compared to the small model. For example, the 9B$/$2B graph visualizes the distribution of the top 10 tokens with the largest \textit{attention\textunderscore weight(Gemma 2\textunderscore 9B)} - \textit{attention\textunderscore weight(Gemma 2\textunderscore 2B)} values. The name tokens (in red), the correct head entity, receive significant additional attention from the larger model.}

\label{fig:attention_compare}
\end{figure*}

Therefore, we can contrast the attention pattern between a large and a small model in the same family to identify the difference in the clue they pay attention to. In Figure \ref{fig:attention_compare}, we subtract the attention weights of the small model from the large model, and visualize the top 10 tokens with the largest attention differences. The graph shows tokens receiving the most ``additional" attention from the large model. It is clear that the correct head entity of the relationship, the ``name" tokens (in red color), often receive the most additional attention\footnote{The date tokens also appear to rank high in attention differences, which is likely due to the fact that there are on average more date tokens than name tokens in the text, so they are counted more frequently in the top 10 tokens. For example, under the LLaMA 3 tokenizer, the name is split into an average of 3.56 tokens, while the date is split into around 7 tokens.}.

Comparing the original model attention in Figure \ref{fig:attention} and the attention difference in Figure \ref{fig:attention_compare}, we can see that while larger models pay more attention to the correct clue in text, the absolute attention weights on the correct clue is still small and biased towards the closer distracting entities. This calls for a method to ``amplify" the attention differences so that the model can focus even more on the correct clue in text.

\section{Method: Augmentation From Contrasting Attention}
\label{method}

We have shown that important clues that are hard to notice in text can be discovered from the attention difference between large and small models. Next, we propose to utilize and amplify these clues by combining with a simple dropout data augmentation method. 

\subsection{Token-Dropout Data Augmentation}

To combat overfitting, token-dropout data augmentation is a simple and effective technique that randomly drops out tokens in a training example \citep{DBLP:conf/emnlp/WeiZ19}. Token-dropout introduces noise to the training data and breaks the model's reliance on spurious co-occurrences in the training examples, helping the model achieve better generalization. A naive token-dropout randomly deletes each token independently with a probability $\alpha$.

\subsection{Augmentation Guided by Elusive Clues}

\begin{figure*}[h]
\includegraphics[scale=0.65, center]{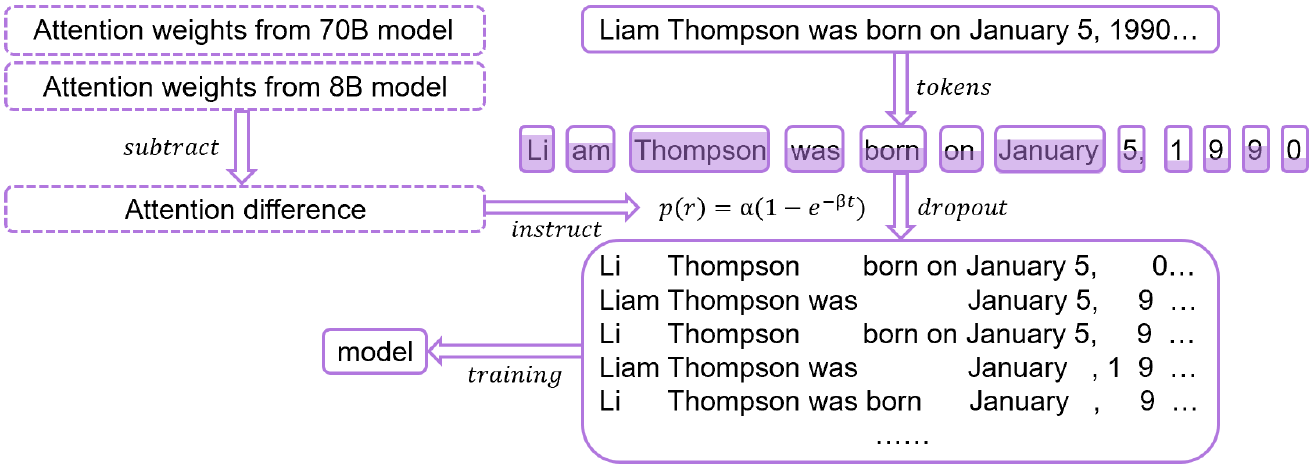}
\caption{Overview of the proposed data augmentation method based on attention difference between large and small models. Color represents retain probability of each token.}
\label{fig:attention-mode-new}
\end{figure*}

Although naive token-dropout mitigates overfitting, it does not solve the long-range dependency learning problem. As each token is dropped out independently, the model still suffers from inadequately small attention to non-obvious and distant information. We propose to use the attention difference between large and small models as a guide to dropout tokens in a more selective way. We first use the attention difference to rank the tokens in the training data, and then dropout tokens with a probability that is inversely proportional to their ranking. In this fashion, the model is encouraged to focus more on the tokens containing important but elusive information, as identified by the attention difference.

We use the following function to calculate dropout probability for each token:
\begin{equation}
    p(r) = \alpha (1 - e^{-\beta r})
    \label{eq:dropout}
\end{equation}
The token with the $r$-th rank (having the $r$-th largest attention difference) will be dropped out with probability $p(r)$. 
% (A graph of the function is shown in Figure \ref{fig:p-function}).
The hyperparameter $\beta$ controls how fast the dropout probability increases with the ranking, and $\alpha$ controls the maximum dropout probability. The tokens with higher attention differences will have lower dropout probabilities, encouraging the model to focus more on these tokens. Figure \ref{fig:attention-mode-new} illustrates the process of the proposed augmentation method.

\section{Results}

\label{result}

\subsection{The Biography Dataset}

We use low-rank adaptation (LoRA) \citep{lora} to facilitate finetuning of models up to 70 billion parameters. As the corpus size is limited, we use a rank of 16 for the LoRA adapters. Adapters are added to all of the model's weights except for the embedding and the output layer.
We finetune models with the Huggingface's transformer library \citep{hf} on NVIDIA 4090 GPUs. 
We experiment with LLaMA 3 \citep{llama3} and Gemma 2 \citep{gemma2} as two families of language models.

For the baselines, we compare the performance of the models after plain finetuning, random (naive) token-dropout, and token-dropout by attention. In addition to random dropout, dropout by attention uses the original attention weights to guide the dropout probabilities, assuming that the model put more attention on tokens it deemed important. Tokens with lower attention weights are dropped out with higher probabilities to enhance the important information, in a similar vein as in \citep{DBLP:journals/ict-express/YuYJK22, DBLP:conf/icmlc2/HailemariamLMA23}. The dropout probabilities are also calculated using Equation \ref{eq:dropout}.

\setlength{\tabcolsep}{0.9mm}

\begin{table}[h]
\centering
\huge
\fontsize{7}{11}\selectfont
\begin{tabular}{lcccccc}
\toprule
& \multicolumn{2}{c}{\small Hyper-} & \multicolumn{4}{c}{\small QA performance} \\
& \multicolumn{2}{c}{\small parameters} & \multicolumn{2}{c}{\small University} & \multicolumn{2}{c}{\small Company} \\
& \small $\bm{\alpha}$ & \small $\bm{\beta}$ & \textbf{\small EM} & \textbf{\small F1} & \textbf{\small EM} & \textbf{\small F1} \\
\midrule
\textit{\small Gemma 2 2B} \\
\textbf{\small Baselines} \\
\small Plain finetuning & \small  - & \small  -  & \small  0.17 &  \small 0.48 &  \small 0.18 & \small  0.21 \\
\small Random token-dropout & \small 0.6 & \small -  & \small 0.07 & \small 0.38 & \small 0.21 & \small 0.23 \\
\small Token-dropout by attention & \small 0.6 & \small 0.05  & \small 0.19 & \small 0.51 & \small 0.23 & \small 0.29 \\
\textbf{\small Ours} \\
\small Token-dropout by attention diff & \small 0.6 & \small 0.03 & \textbf{\small 0.25} & \textbf{\small 0.56} & \textbf{\small 0.32} & \textbf{\small 0.36} \\
\midrule
\textit{\small Gemma 2 9B} \\
\textbf{\small Baselines} \\
\small Plain finetuning & \small - & \small -  & \small 0.61 & \small 0.78 & \small 0.63 & \small 0.64 \\
\small Random token-dropout & \small 0.7 & \small -  & \small 0.52 & \small 0.73 & \small 0.51 & \small 0.57 \\
\small Token-dropout by attention & \small 0.6 & \small 0.05  & \small 0.49 & \small 0.62 & \small 0.44 & \small 0.47 \\
\textbf{\small Ours} \\
\small Token-dropout by attention diff & \small 0.6 & \small 0.03 & \textbf{\small 0.84} & \textbf{\small 0.92} & \textbf{\small 0.90} & \textbf{\small 0.92} \\
\midrule
\textit{\small LLaMA 3 8B} \\
\textbf{\small Baselines} \\
\small Plain finetuning & \small - & \small -  & \textbf{\small 0.30} & \small 0.55 & \small 0.17 & \small 0.21 \\
\small Random token-dropout & \small 0.6 & \small -  & \small 0.11 & \small 0.49 & \small 0.24 & \small 0.29 \\
\small Token-dropout by attention & \small 0.6 & \small 0.05  & \small 0.24 & \small 0.62 & \small 0.21 & \small 0.28 \\
\textbf{\small Ours} \\
\small Token-dropout by attention diff & \small 0.7 & \small 0.05  & \small 0.29 & \textbf{\small 0.64} & \textbf{\small 0.42} & \textbf{\small 0.53} \\
\midrule
\textit{\small LLaMA 3 70B} \\
\textbf{\small Baselines} \\
\small Plain finetuning & \small - & \small -  & \small 0.42 & \small 0.69 & \small 0.66 & \small 0.67 \\
\small Random token-dropout & \small 0.6 & \small -  & \small 0.71 & \small 0.86 & \small 0.71 & \small 0.78 \\
\small Token-dropout by attention & \small 0.7 & \small 0.05  & \small 0.51 & \small 0.75 & \small 0.61 & \small 0.68 \\
\textbf{\small Ours} \\
\small Token-dropout by attention diff & \small 0.7 & \small 0.01 & \textbf{\small 0.90} & \textbf{\small 0.96} & \textbf{\small 0.96} & \textbf{\small 0.96} \\
\bottomrule
\end{tabular}
\caption{QA performance after continual pretraining on the biography corpus.
Data augmentation based on attention difference significantly outperforms other data augmentation methods, for both small and large models.}
\label{tab:syn_results}
\end{table}

For each experiment, we trained the model from 10 to 30 epochs with learning rates in [5e-5, 1e-3] and selected the model with the best performance. For the augmentation-based methods, we also searched for the best hyperparameters $\alpha$ and $\beta$ individually for each method. Interestingly, the best hyperparameters for the dropout probabilities happen to be similar for different models and augmentation methods. For each of the augmentation methods, we generate 10 augmented versions of each training example and combine them with the original examples.

Results in Table \ref{tab:syn_results} show that the proposed token-dropout augmentation based on attention difference significantly outperforms other data augmentation methods. We report QA accuracy on the ``university" and the ``company" fields as the models have poor performance on these fields under plain finetuning (Table \ref{tab:plain-finetuning}). We report exact match (EM) accuracy and normalized word-level F1 scores. We can see that while random dropout and dropout by attention improve performance over no data augmentation, our method achieves much more significant improvement. 
We also collected the results of other information from models trained in our method and accuracy increased across models.
This proves that contrasting attention of large and small language models indeed finds important but elusive clues in text effectively, and amplifying these clues in the input has immediate positive effects on the model's memorization efficiency even for the 70B model.

\subsection{Real-World Dataset}

Aside from the biography dataset, we also evaluate the proposed method on Wikipedia text to verify if the method helps knowledge learning on general text. Specifically, we evaluate on the Paragraph-Level Wikipedia Question-Answering dataset \citep{du-cardie-2018-harvesting}. We first perform continual pretraining on the Wikipedia text paragraphs (included in the dataset), then evaluate the model's performance on the question-answering data\footnote{This is the ``closed-book" setting where the model is not allowed to look at the original Wikipedia passage during question answering. It tests the model's ability to memorize factual knowledge during the continual pretraining phase.}.
The questions are specifically designed to incorporate coreference dependencies that span multiple sentences in a paragraph, making it a challenging task that tests the model's ability to learn and memorize complex factual associations.

An example of Wikipedia text from the dataset is:

\begin{quote}
    \textit{The 2005 edition of the International ISBN Agency's official manual describes how the 13-digit ISBN check digit is calculated. The ISBN-13 check digit, which is the last digit of the ISBN, must range from 0 to 9 and must be such that the sum of all the thirteen digits, each multiplied by its (integer) weight, alternating between 1 and 3, is a multiple of 10.}
\end{quote}

An example of the question from the dataset is as follows:

\begin{quote}
\textit{Question: How many digits does the ISBN have?
\\
Answer: 13}
\end{quote}

\setlength{\tabcolsep}{1.5mm}

\begin{table}[h]
\centering
\huge
\fontsize{7}{11}\selectfont
\begin{tabular}{lcccc}
\toprule
& \multicolumn{2}{c}{\small Hyper-} & \multicolumn{2}{c}{\small QA} \\
& \multicolumn{2}{c}{\small parameters} & \multicolumn{2}{c}{\small performance} \\
& \small $\bm{\alpha}$ & \small $\bm{\beta}$ & \textbf{\small EM} & \textbf{\small F1}  \\
\midrule
\textit{\small Gemma 2 2B} \\
\textbf{\small Baselines} \\
\small Plain finetuning & \small - & \small -  & \small 0.126 & \small 0.215 \\
\small Random token-dropout & \small 0.7 & \small -  & \small 0.12 & \small 0.223 \\
\small Token-dropout by attention & \small 0.7 & \small 0.005  & \small 0.145 & \small 0.249 \\
\textbf{\small Ours} \\
\small Token-dropout by attention diff & \small 0.7 & \small 0.005  & \textbf{\small 0.156} & \textbf{\small 0.256} \\
\midrule
\textit{\small Gemma 2 9B} \\
\textbf{\small Baselines} \\
\small Plain finetuning & \small - & \small -  & \small 0.186 & \small 0.287 \\
\small Random token-dropout & \small 0.7 & \small -  & \small 0.198 & \small 0.314 \\
\small Token-dropout by attention & \small 0.7 & \small 0.005  & \small 0.205 & \small 0.315 \\
\textbf{\small Ours} \\
\small Token-dropout by attention diff & \small 0.7 & \small 0.005  & \textbf{\small 0.231} & \textbf{\small 0.334} \\
\midrule
\textit{\small LLaMA 3 8B} \\
\textbf{\small Baselines} \\
\small Plain finetuning & \small - & \small -  & \small 0.146 & \small 0.228 \\
\small Random token-dropout & \small 0.7 & \small -  & \small 0.067 & \small 0.159 \\
\small Token-dropout by attention & \small 0.7 & \small 0.005  & \small 0.134 & \small 0.239 \\
\textbf{\small Ours} \\
\small Token-dropout by attention diff & \small 0.7 & \small 0.03  & \textbf{\small 0.172} & \textbf{\small 0.263} \\
\midrule
\textit{\small LLaMA 3 70B} \\
\textbf{\small Baselines} \\
\small Plain finetuning & \small - & \small -  & \small 0.179 & \small 0.282 \\
\small Random token-dropout & \small 0.7 & \small -  & \small 0.187 & \small 0.307 \\
\small Token-dropout by attention & \small 0.7 & \small 0.005  & \small 0.190 & \small 0.288 \\
\textbf{\small Ours} \\
\small Token-dropout by attention diff & \small 0.7 & \small 0.005  & \textbf{\small 0.212} & \textbf{\small 0.308} \\
\bottomrule
\end{tabular}
\caption{QA performance after continual pretraining on the Wikipedia corpus.
Data augmentation based on attention difference outperforms other data augmentation methods.}
\label{tab:wiki_results}
\end{table}

Results in Table \ref{tab:wiki_results} show that the proposed method also improves knowledge learning from the Wikipedia text. Unlike naive data augmentation, our method improves the model's memorization efficiency by selectively amplifying difficult and elusive clues. This shows that enhancing the model's focus on important but elusive information is a crucial factor in improving knowledge learning efficiency, and our method is generally applicable to different kinds of text.

\section{Conclusion}

Efficiency of learning factual knowledge in not only crucial for pretraining, but also important for effective continual and lifelong learning in language models. Due to the overfitting and long-range dependency problem, even performant language models can struggle to learn and memorize factual knowledge from limited data. In this work, we show that one of the key factors to improving the model's learning, finding the ``elusive" but important clues in text, is already embedded in the model's attention weights. However, such clues are hard to discover by the model itself due to the model's bias towards short-range contexts, but clearly manifests themselves when contrasting the attention between a larger and a smaller model. Based on this discovery, we propose a simple yet effective data augmentation method that leverages the attention difference to guide the dropout of tokens in the input. Our method significantly improves the model's performance in memorizing factual knowledge, and is shown to be effective for different corpora and models.

\section{Acknowledgements}

This work was supported by the Noncommunicable Chronic Diseases-National Science and Technology Major Project (Grant No. 2023ZD0506501).

% \clearpage

\begin{small}
\bibliography{aaai25}
\end{small}

\clearpage

\end{document}